\documentclass[10pt,twocolumn,letterpaper]{article}

\usepackage[pagenumbers]{cvpr} 

\usepackage{colortbl}
\usepackage{xcolor}
\usepackage{booktabs}
\usepackage{array}

\usepackage{tikz}

\definecolor{titleblue}{HTML}{1A3A5C}
\definecolor{sectionblue}{HTML}{2C5F8A}
\definecolor{lightgray}{HTML}{F0F0F0}

\definecolor{cvprblue}{rgb}{0.21,0.49,0.74}
\usepackage[pagebackref,breaklinks,colorlinks,allcolors=cvprblue]{hyperref}

\def\paperID{45} 
\def\confName{CVPR}
\def\confYear{2026}

\title{Evaluation Cards for XAI Metrics}

\author{Rokas Gipiškis$^{1,2}$  \quad Olga Kurasova$^{1}$\\
$^{1}$Vilnius University \quad $^{2}$AI Standards Lab\\
{\tt\small rokas.gipiskis@mif.vu.lt}
}

\begin{document}
\maketitle
\AddToShipoutPicture*{%
  \AtPageLowerLeft{%
    \put(72, 20){\parbox{\textwidth}{\footnotesize Accepted at the 5th Explainable AI for Computer Vision (XAI4CV) Workshop, CVPR 2026 (non-archival).}}%
  }%
}

\begin{abstract}
The evaluation of explainable AI (XAI) methods is affected by a lack of standardization. Metrics are inconsistently defined, incompletely reported, and rarely validated against common baselines. In this paper, we identify transparency of evaluation reporting as a central, under-addressed problem. We propose the XAI Evaluation Card, a documentation template analogous to model cards, designed to accompany any study that introduces an XAI evaluation metric. The card covers explicit declaration of target properties, grounding levels, metric assumptions, validation evidence, gaming risks, and known failure cases. We argue that adopting this template as a community norm would reduce evaluation fragmentation, support meta-analysis, and improve accountability in XAI research.
\end{abstract}    
\section{Introduction}
\label{sec:intro}

As AI systems are deployed in high-stakes settings, explainable AI (XAI) has become essential for accountability and transparency. However, the difficult part is not just explaining model predictions, but rigorously evaluating those explanations. A growing body of literature has attempted to systematize XAI evaluation metrics, but the field remains fragmented. Metrics are introduced without clear mappings to the properties they measure, and reported without contextual information needed for reproducibility.

This paper focuses on standardizing documentation for XAI evaluation through the XAI Evaluation Card. Drawing on a meta-review of surveys published between 2021 and 2025, we argue that the absence of structured reporting is a primary addressable cause of many of the field's evaluation problems, and that a card-based template, analogous to model cards \cite{mitchell2019model} and datasheets \cite{gebru2021datasheets}, can address them.

Our contribution is a concrete template with four structured sections covering identity, scope, validation, and relationships. We discuss how the template maps onto recurring gaps identified across the surveyed literature, and how its adoption as a community norm could improve reproducibility and comparability in XAI evaluation.
\section{Background and Motivation}
\label{sec:formatting}

Across eleven surveys covering XAI evaluation metrics published between 2021 and 2025, three main problems appear. First, the field lacks terminological consensus. The same metric names appear under different definitions across papers, and the same underlying properties can be named differently by different research groups. Second, evaluation practice is heavily skewed toward functionally-grounded proxy tasks while human-grounded and application-grounded methods remain underrepresented despite being more informative about real-world utility. Third, implementation availability is inconsistently reported, with some proposed metrics remaining theoretical definitions without accompanying code.

\subsection{Meta-review}

Terminological fragmentation is documented most extensively by \citet{pawlicki2024evaluating}, who identify close to a hundred XAI metrics and note that several distinct papers use identical metric names for different operationalizations, while papers measuring the same underlying property use different terminology for it. \citet{kadir2023evaluation} observe the resulting difficulty in benchmarking and comparison across XAI evaluation methods. \citet{nauta2023anecdotal} also note standardization-related disagreements, and attempt a systematic response, reviewing 29 quantitative evaluation methods across 12 evaluation properties and arguing that explainability must be treated as a multi-faceted concept evaluated along multiple dimensions simultaneously. 

The skew toward functionally-grounded evaluation is noted by \citet{lopes2022xai}, whose taxonomy distinguishes human-centered from computer-centered evaluations, and by \citet{mangold2025design}, who argue that current evaluation processes are often too technical and insufficiently focused on human users. \citet{mohseni2021multidisciplinary} find that human-AI task performance dominates the evaluation literature, while mental model alignment and user trust are rarely operationalized. \citet{sisk2022analyzing} similarly highlight a lack of consensus on how explanation reliability and validity should be assessed from a human perspective.

Implementation gaps are highlighted by \citet{coroama2022evaluation}, who check reviewed papers for the presence or absence of an accompanying implementation, finding that many proposed metrics remain theoretical constructs without practical instantiation. \citet{banerjee2022methods} similarly note that quantitative metrics depend heavily on the type of machine learning problem and model used, which compounds the difficulty of providing general-purpose implementations. \citet{zhou2021evaluating} find that quantitative metrics for model-based and example-based explanations largely measure simplicity, while metrics for attribution-based explanations target soundness. The most recent survey, \citet{dembinsky2025unifying}, attempts to unify the field through the VXAI framework, consolidating 362 publications into 41 functionally-similar metric groups and proposing a three-dimensional categorization scheme. However, this framework focuses on taxonomy rather than documentation standards, and does not address the reporting gaps we identify.

\subsection{The Evaluation Problem in XAI}

A fundamental distinction in XAI is between properties (conceptual qualities such as fidelity, robustness, or clarity) and metrics, which operationalize these properties into measurable quantities. This distinction is important because the mapping from properties to metrics is neither unique nor exact. Multiple metrics may target the same property while giving different conclusions, and a metric's validity is often context-dependent, varying with model architecture, data modality, and explanation scope.

Across eleven reviewed surveys, five recurring evaluation problems were identified: (1) metrics are introduced without declaring which properties they target; (2) results are reported without specifying the evaluation context in which they are valid; (3) few metrics include sensitivity or stability analysis in their original proposal; (4) metric disagreements are rarely acknowledged and interpreted; and (5) implementation availability is inconsistent, with many metrics remaining at the level of theoretical definitions.

These problems are not independent. The absence of declared target properties makes disagreement handling difficult, and the lack of contextual reporting prevents meaningful replication. The XAI Evaluation Card is designed to address all five, as detailed in Section \ref{evalcard}.

\subsection{Related Documentation Frameworks}

The idea of structured documentation for AI artifacts is already established in other areas. Model cards \cite{mitchell2019model} standardize reporting of model performance. Datasheets for Datasets \cite{gebru2021datasheets} document dataset origin, collection procedures, and intended uses. To our knowledge, no equivalent documentation standard exists specifically for XAI evaluation metrics, despite the fact that these metric results are currently used as evidence for claims about explanation quality.

\section{The XAI Evaluation Card}
\label{evalcard}

We propose the XAI Evaluation Card as a standardized supplement to any study that proposes a new XAI evaluation metric. The card is organized into four sections, each addressing a different class of documentation gap identified in the meta-review. A filled out example can be found in Appendix \ref{appendix}.

\begin{table*}[ht]
\centering
\small
\setlength{\tabcolsep}{6pt}
\renewcommand{\arraystretch}{1.25}
\begin{tabular}{>{\bfseries\raggedright\arraybackslash}p{2.8cm} p{11.4cm}}

\multicolumn{2}{c}{\cellcolor{titleblue}\textcolor{white}{\textbf{\normalsize XAI Evaluation Card}}} \\[2pt]

\multicolumn{2}{l}{\cellcolor{sectionblue}\textcolor{white}{\textbf{I.\quad Identity}}} \\
\rowcolor{lightgray} Metric Name
  & Unique, descriptive name for the evaluation metric. \\
Target Property / Properties
  & List all explainability properties this metric operationalizes (e.g.,
    \textit{fidelity}, \textit{robustness}, \textit{clarity}), with references
    to definitions used. \\
\rowcolor{lightgray} Grounding Level
  & One or more of: \textit{functionally-grounded} / \textit{human-grounded} /
    \textit{application-grounded} \citep{doshi2017towards}. \\[4pt]

\multicolumn{2}{l}{\cellcolor{sectionblue}\textcolor{white}{\textbf{II.\quad Scope and Context}}} \\
\rowcolor{lightgray} Evaluation Context
  & Model architecture, data modality, and explanation scope
    (\textit{local} / \textit{global}) under which results are reported. \\
Assumptions
  & All assumptions required by the metric (e.g., feature independence,
    locality, linearity, calibrated probabilities, meaningful baselines). \\[4pt]

\multicolumn{2}{l}{\cellcolor{sectionblue}\textcolor{white}{\textbf{III.\quad Implementation and Validation}}} \\
\rowcolor{lightgray} Implementation Available?
  & Yes\,/\,No. If yes, provide URL or repository reference. \\
Validation Evidence
  & Summary of sensitivity analysis, stability analysis, and correlation with
    related metrics. Report computational cost where relevant. \\
\rowcolor{lightgray} Gaming Risk
  & How a method could achieve a high score on this metric without improving
    the target explainability property. \\
Known Failure Cases
  & Conditions under which the metric is known to fail or produce misleading
    results. \\[4pt]

\multicolumn{2}{l}{\cellcolor{sectionblue}\textcolor{white}{\textbf{IV.\quad Relationships and Limitations}}} \\
\rowcolor{lightgray} Relationship to Other Metrics
  & Metrics targeting the same property. Known agreements or disagreements
    in results. \\
Disagreement Handling
  & If this metric conflicts with others reported, state which property is
    prioritised for the target deployment scenario and why. \\
\rowcolor{lightgray} Limitations
  & Main limitations as an operationalization of the target property. Note
    contexts where the metric should not be used. \\

\end{tabular}
\caption{XAI Evaluation Card template. Fields marked N/A require a brief justification.}
\label{tab:eval-card}
\end{table*}

\subsection{Identity}

The identity section requires authors to name the metric, list all explainability properties it operationalizes (with explicit references to property definitions), and declare its grounding level following the framework of \citet{doshi2017towards}: functionally-grounded (proxy tasks), human-grounded (user studies), or application-grounded (domain experts in deployment settings). This addresses the widespread conflation of properties and metrics, and the lack of declared evaluation grounding observed across the surveyed literature. Explicit grounding declarations prevent a common failure mode: drawing human-centered conclusions from purely technical metrics.

\subsection{Scope and Context}

The scope section requires a description of the evaluation context (model architecture, data modality, and whether evaluation is local or global) and an explicit enumeration of all assumptions made by the metric, such as feature independence, locality, or the availability of meaningful baselines. \citet{coroama2022evaluation} highlight that metric validity is highly context-dependent, yet contextual information is routinely omitted from publications. Without it, reported metric scores are not interpretable and cannot be meaningfully compared across studies. This section operationalizes the principle that metric values should never be reported in isolation.

\subsection{Implementation and Validation}

The validation section asks authors to (a) indicate whether an implementation is available and provide a link, (b) summarize validation evidence including sensitivity analysis, stability analysis, and correlation with related metrics, (c) describe gaming risk (how a method could score highly on this metric without actually improving the target property) and (d) document known failure cases. These requirements respond directly to findings that some proposed metrics lack implementations and that validation beyond the original development context is rare. The gaming risk field is particularly important as it highlights a class of validity threat that is rarely made explicit in evaluation papers.

\subsection{Relationships and Limitations}

The final section places the metric in its broader evaluation ecosystem. Authors should list metrics targeting the same property and note any known agreements or disagreements. When metrics diverge, they should state which property is prioritized and why, for the target deployment scenario. A limitations field requires listing conditions under which the metric should not be used. \citet{pawlicki2024evaluating} find a large diversity of metrics without consensus on their properties, making inter-metric relationships a critical missing element of most publications. This section makes those relationships explicit and searchable, directly supporting meta-analysis.

\section{Discussion}
\label{sec:discussion}

\subsection{Relevance to Evaluation Practice}
Each field in the XAI Evaluation Card maps onto a documented failure mode in the literature. Cards can be completed as a supplementary table or appendix, imposing minimal overhead while creating a structured, machine-readable record of evaluation decisions. Requiring cards as part of peer review would shift evaluation norms toward greater rigor.

The card is explicitly non-prescriptive. It does not mandate any particular grounding level, metric, or validation procedure. Fields that are not applicable may be marked N/A with justification. This design choice reflects the diversity of XAI evaluation contexts, from post-hoc attribution methods on tabular data to global explanations of deep vision models, and avoids imposing a single evaluation paradigm.

\subsection{Connections to Model Cards and Datasheets}

The XAI Evaluation Card is intentionally analogous to model cards \cite{mitchell2019model} and datasheets \cite{gebru2021datasheets}, which have achieved significant community adoption. The card differs from these in its focus: where model cards document what a model does and for whom, XAI Evaluation Cards document how the quality of an explanation is being measured and under what conditions that measurement is valid. 

\subsection{Adoption}

A well-designed card can fail without wider adoption, and the documentation burden it adds should be justified to authors. We see three ways to lower this barrier and incentivize adoption. First, we propose integration with the reproducibility checklists already used by major venues, so the XAI Evaluation Card is completed alongside existing reporting requirements. Second, we suggest using lightweight templates and schemas to lower the authoring cost. This could include Markdown and LaTeX templates, a defined JSON schema so completed cards can be parsed and indexed by future efforts, and LLM-assisted drafting that extracts factual fields (metric name, modality, grounding level, implementation link) from the paper text and leaves judgement-heavy fields (target properties, gaming risk, known failure cases, relationships) to the author. Third, we recommend gradual venue-level adoption, beginning with workshops where the community is concentrated and feedback loops are short, and expanding to other venues once value is demonstrated. Within participating venues, reviewer checklists asking whether a card is provided and whether key fields are filled with sufficient detail could serve as a soft enforcement mechanism. Mandating XAI Evaluation Cards at the reviewer level is probably the strongest enforcement mechanism, but the steps above make voluntary uptake plausible before any such mandate.

\subsection{Limitations}

The card template is a minimal viable standard. It does not resolve deeper disagreements about what properties XAI explanations should satisfy, nor does it provide a formal ontology or taxonomy for aligning differently-named concepts across studies (e.g., whether "faithfulness" in one paper corresponds to "fidelity" in another). Future work could extend the card with formal property references and machine-readable schemas to support automated meta-analysis. 

\section{Conclusion}
\label{sec:conclusion}

XAI evaluation is a critical yet fragmented field. Motivated by a meta-review of eleven recent surveys, we have identified transparency of evaluation reporting as a central, under-addressed challenge. We propose the XAI Evaluation Card, a four-section structured documentation template covering metric identity, evaluation scope and context, implementation and validation, and inter-metric relationships. We argue that adopting this template as a community norm would reduce evaluation fragmentation, enable meaningful comparison across studies, and improve the accountability of empirical claims about explanation quality.

{
    \small
    \bibliographystyle{ieeenat_fullname}
    \bibliography{main}
}

\appendix
\clearpage
\setcounter{page}{1}
\maketitlesupplementary

\clearpage
\onecolumn 
\section{Example of a Filled Out Evaluation Card}
\label{appendix}

\begin{table*}[ht]
\centering
\small
\setlength{\tabcolsep}{6pt}
\renewcommand{\arraystretch}{1.25}
\begin{tabular}{>{\bfseries\raggedright\arraybackslash}p{2.8cm} p{11.4cm}}

\multicolumn{2}{c}{\cellcolor{titleblue}\textcolor{white}{\textbf{\normalsize XAI Evaluation Card}}} \\[2pt]

\multicolumn{2}{l}{\cellcolor{sectionblue}\textcolor{white}{\textbf{I.\quad Identity}}} \\
\rowcolor{lightgray} Metric Name
  & Deletion Area Under the Curve (Deletion AUC / DAUC) \cite{petsiuk2018rise} \\
Target Property / Properties
  & Faithfulness / Fidelity. It operationalizes this by measuring if the features identified as "important" by the explanation are necessary for the model to maintain its predictive confidence. \\
\rowcolor{lightgray} Grounding Level
  & Functionally-grounded (proxy task). \\[4pt]

\multicolumn{2}{l}{\cellcolor{sectionblue}\textcolor{white}{\textbf{II.\quad Scope and Context}}} \\
\rowcolor{lightgray} Evaluation Context
  & Applied to local feature attributions (e.g., saliency maps) across vision (both classification and segmentation tasks \cite{gipivskis2024explainable}) and NLP modalities. Requires a model with probability or logit outputs. \\
Assumptions
  & Assumes that iteratively masking top-rated features will degrade model performance if the explanation is faithful. Assumes the chosen baseline/imputation method (e.g., replacing pixels with zeros, mean values, or blurring) is meaningful and does not artificially break the model. \\[4pt]

\multicolumn{2}{l}{\cellcolor{sectionblue}\textcolor{white}{\textbf{III.\quad Implementation and Validation}}} \\
\rowcolor{lightgray} Implementation Available?
  & Yes. https://github.com/eclique/RISE/blob/master/evaluation.py. \\
Validation Evidence
  & Empirical studies show DAUC is highly sensitive to the choice of the baseline/imputation value (e.g., zero-masking vs. generative inpainting). It carries a moderate-to-high computational cost, requiring multiple forward passes per instance as features are incrementally removed. \\
\rowcolor{lightgray} Gaming Risk
  & An explanation method could achieve a high DAUC score by intentionally selecting features that, when masked, create severe out-of-distribution (OOD) artifacts. The model's confidence drops because the input looks unnatural (like adversarial noise), not because the true explanatory features were removed. \\
Known Failure Cases
  & Can produce misleading results when features are highly correlated. The model might rely on a redundant, unmasked feature, making the DAUC score artificially low despite a good explanation. \\[4pt]

\multicolumn{2}{l}{\cellcolor{sectionblue}\textcolor{white}{\textbf{IV.\quad Relationships and Limitations}}} \\
\rowcolor{lightgray} Relationship to Other Metrics
  & Conceptually similar to comprehensiveness (used in NLP). Often paired with the Insertion AUC metric. May disagree with Faithfulness Correlation if the model's response to feature removal is highly non-linear. \\
Disagreement Handling
  & If DAUC conflicts with Insertion AUC, DAUC should be prioritized if the deployment scenario strictly requires identifying the features that are necessary for the model to work (e.g., safety auditing for failure modes). \\
\rowcolor{lightgray} Limitations
  & The main limitation is the OOD problem. The metric might evaluate the model's robustness to missing data rather than the explanation's true fidelity. Should not be used in isolation without an insertion or OOD-compensated baseline. \\

\end{tabular}
\caption{Example of a filled out XAI Evaluation Card for Deletion Area Under the Curve (Deletion AUC / DAUC), a popular metric used to evaluate feature attribution methods.}
\label{tab:eval-card}
\end{table*}
\appendix

\end{document}